\documentclass[preprint,12pt,authoryear]{elsarticle}

\usepackage{amssymb}
\usepackage{adjustbox}
\usepackage{float}
\usepackage{color,soul}
\usepackage{listings}
\usepackage{xcolor}
\usepackage{multirow}
\usepackage{amsmath}
\usepackage{graphicx}

\restylefloat{table}

\journal{Information Fusion}

\begin{document}

\begin{frontmatter}




\title{Semantic XAI for contextualized demand forecasting explanations}

\author[1,2,3]{Jo\v{z}e M. Ro\v{z}anec\corref{cor1}}
\ead{joze.rozanec@ijs.si}
\cortext[cor1]{Corresponding author.}

\author[1]{Dunja Mladeni\'{c}}
\ead{dunja.mladenic@ijs.si}

\address[1]{Jo\v{z}ef Stefan Institute, Jamova 39, 1000 Ljubljana, Slovenia}
\address[2]{Qlector d.o.o., Rov\v{s}nikova 7, 1000 Ljubljana, Slovenia}
\address[3]{Jo\v{z}ef Stefan International Postgraduate School, Jamova 39, 1000 Ljubljana, Slovenia}


%
%

\begin{abstract}
The paper proposes a novel architecture for explainable AI based on semantic technologies and AI. We tailor the architecture for the domain of demand forecasting and validate it on a real-world case study. The provided explanations combine concepts describing features relevant to a particular forecast, related media events, and metadata regarding external datasets of interest. The knowledge graph provides concepts that convey feature information at a higher abstraction level. By using them, explanations do not expose sensitive details regarding the demand forecasting models. The explanations also emphasize actionable dimensions where suitable. We link domain knowledge, forecasted values, and forecast explanations in a Knowledge Graph. The ontology and dataset we developed for this use case are publicly available for further research.
\end{abstract}


\begin{highlights}
\item We developed a novel architecture for XAI based on semantic technologies and AI.
\item We validated the architecture on a real-world case study regarding demand forecasting.
\item Explanations avoid exposing AI model features while informing about them.
\item We enrich explanations with related real-world events reported by news media.
\item Explanations recommend new datasets to enhance future demand forecasting models.
\end{highlights}

\begin{keyword}
Explainable AI \sep Knowledge Graph \sep Demand Forecasting \sep Smart Manufacturing


\end{keyword}

\end{frontmatter}


\section{Introduction}\label{INTRODUCTION}
The increasing digitalization of manufacturing in the context of Industry 4.0 provides a growing amount of data describing assets and operations. This data increases the transparency of production processes, accelerates the information flow through the company, and is a valuable asset to build predictive models. Demand forecasting is an essential component for any manufacturing company. A growing body of research has addressed techniques that identify demand-type and develop statistical and AI models to predict future demand. In critical operations, such as demand forecasting, it is crucial to provide a forecast and uncertainty and convey some explanations to the user. In this way, the user can make an informed decision, which may have far-reaching consequences, and avoid costly mistakes. Such explanations enhance the AI models, increase trust in the system, and help identify errors and performance issues.

Researchers developed several approaches related to AI models' explainability, dividing the models into glass-box (the AI algorithm can explain its prediction) and black-box (we need another algorithm to obtain an explanation of a model). Such explanations can be provided on a global level (the forecasting model in general) or at a local level (for every prediction instance). Some authors also envisioned that the inclusion of semantic technologies could be used to determine the semantic closeness of concepts encoded in data features or integrate knowledge graph embeddings to a forecasting model to produce explanations along with the forecasts (\cite{gaur2020semantics,panigutti2020doctor}).

Explanation of an AI model should provide cues on factors that influenced a forecast, provide means to discern relevant context, supporting information, and emphasize actionable aspects to assist decision-making. In our research, we focus on local-level explanations so that planners can still weight cues provided for each case. We develop a domain-specific ontology and knowledge graph to provide domain knowledge and context for creating explanations. The knowledge graph allows us to identify which cues correspond to semantically close concepts and what variables people may influence, support information from the dataset and external systems. In particular, we integrate a Media Event Retrieval System to match provided cues to relevant news and identify other concepts that may be relevant to the case. The system then proposes to the user such concepts and potential datasets that could be used to enrich the existing data.

The main scientific contributions of the presented research are the following:
\begin{enumerate}
\item a modular architecture to provide demand forecasting explanations, which consider semantic proximity, highlight factors that can be influenced by the user, relevant context, and opportunities for data enrichment;
\item a unified ontology to capture domain knowledge regarding demand forecasting, events, and other elements relevant to the explanations;
\item the architecture implementation and validation on a real-world use case in manufacturing;
\item publicly available dataset containing: features relevant to forecasts, retrieved media events, metadata regarding external datasets, and a subset of annotated media events and external datasets metadata. Such a dataset can be a valuable resource for further research on identifying, ranking relevant, and summarizing entries to explain forecasts better.
\end{enumerate}

To evaluate the overall architecture, we manually annotate a set of 528 media events, 401 datasets metadata entries, and 168 forecast explanations and assess their goodness based on metrics and criteria described in Section \ref{EVALUATION}.

The rest of this paper is structured as follows: Section \ref{RELATED-WORK} presents related work, Section \ref{ARCHITECTURE} describes the proposed architecture in the context of demand forecasting, Section \ref{ONTOLOGY} describes the ontology we used to construct the knowledge graph, Section \ref{USE-CASE} describes the use case for which we implemented the architecture, Section \ref{EVALUATION} provides the results we obtained. Finally, in Section \ref{CONCLUSION}, we provide our conclusions and outline future work ideas.

\section{Related Work}\label{RELATED-WORK}

Demand forecasting is a crucial component for any manufacturing company. Accurate forecasts translate to production planning, influencing raw materials purchases, production line schedules, and inventory. AI models can play a pivotal role in enhancing demand forecasting (\cite{lolli2017single,bergman2017bayesian,babai2020empirical}). Extensive research has been performed, considering different products and demand types (\cite{gutierrez2008lumpy,kourentzes2013intermittent,carbonneau2008application}). Such models can consider a wide range of data sources and learn non-trivial patterns and correlations that may not be easy to find based only on the human experience. Demand forecasts with their associated uncertainty do not provide enough ground for planners' decision making. Planners require to gain insight into the reasons driving such a forecast and context to evaluate the forecast's soundness.

\subsection{XAI: black-box models and semantic technologies}\label{RELATED-WORK-BBM}
Human-understandable explanations in the demand forecasting domain are essential as they help the planner to (I) understand which of the model attributes are strong drivers in determining the expected demand, (II) identify some kind of bias in the model, and (III) understand how the model can be improved.

There are two approaches to issue explanations regarding a model prediction: glass-box and black-box. We consider that given the velocity of research on AI models, it is preferable to treat models as black-boxes. By doing so, we decouple the forecast and explainability dimension and provide greater flexibility in choosing the best approach for each of them.

Researchers developed multiple approaches to provide black-box explanations of forecasting models. Among most frequently cited we find LIME (\cite{ribeiro2016should}) and its variants (e.g.: k-LIME (\cite{hall2017machine}), DLIME (\cite{zafar2019dlime}), and LIMEtree (\cite{sokol2020limetree})), Anchors (\cite{ribeiro2018anchors}),  Local Foil Trees (\cite{van2018contrastive}), or LoRE (\cite{guidotti2018local}). These approaches build surrogate models for each prediction sample, learning the reference model's behavior on the particular case of interest by introducing perturbations to the feature vector variables.

Good explanations should convey meaningful information, resemble a logic explanation (\cite{pedreschi2018open}), target a specific user profile (\cite{samek2019towards}), focus on actionability (\cite{verma2020counterfactual}), and if possible, provide some counterfactuals. When focusing on a specific user profile, it is important to adapt the language and content of relevant information that should or can be shown to it. The actionable dimension relates to the ability to distinguish mutable features that can be influenced by the user.

Multiple authors envisioned the usage of semantic technologies in the explainability domain. Doctor XAI (\cite{panigutti2020doctor}) develops an agnostic XAI technique for ontology-linked data classification by training a surrogate model and extracting rules from it. \cite{gaur2020semantics} make use of a Knowledge Graph to feed deep learning models with it to enhance their explainability. \cite{samek2019towards} envision Knowledge Graphs can be used to compact large tree models by combining nodes into unique probabilistic concepts. In addition to the opportunities mentioned above, we consider semantic technologies shall (A) provide background knowledge which can be leveraged to provide semantic meaning to dataset features, (B) inform their characteristics (e.g., valid ranges, if they are mutable or immutable if the user can influence their values), or (C) issue explanations with adequate language and context, considering if certain information shall be or not informed to the user.

A modular architecture that decouples forecasts from their explanations is required to provide the planers with adequate explanations regarding machine demand forecasts. Such architecture has the advantage of allowing to iterate both components at their own pace. Black-box explainability models mostly lack domain knowledge and inform relevant forecasting features without a broader context and a more profound interpretation. A third module can provide such context and interpretation, taking into account black-box explanations, domain-knowledge encoded in an ontology and instantiated in a Knowledge Graph, to create a better explanation for the end-user.

\subsection{Domain specific ontologies}\label{RELATED-WORK-DSO}
Demand Forecasting is an essential activity in manufacturing, and manufacturing domain ontologies provide some concepts related to it. E.g., \cite{ameri2006upper} developed the Manufacturing Service Description Language ontology as a formal representation of manufacturing services, and \cite{lemaignan2006mason} described knowledge regarding individual operations in MASON.

Since our architecture links concepts from datasets used to train the AI models that issue the forecasts, this should also be encoded in the unified ontology. To promote the reuse of concepts, following the MIREOT principle (\cite{courtot2011mireot}), we analyzed ontologies related to the artificial intelligence domain. \cite{cannataro2003data} developed the DAMON (Data Mining Ontology for Grid Programming), which provides a reference model for data mining tasks, methodologies, and available software. A heavyweight ontology was developed by \cite{panov2008ontodm,panov2014ontology}, who provide means to represent data mining entities, inductive queries, and data mining scenarios. \cite{diamantini2009kddonto} developed KDDONTO, focusing on data mining algorithms discovery.

The explanation must provide information regarding features that are most influential to a forecast. This view shall be complemented with context, describing events that illustrate possible reasons behind the values captured in the features, and point to similar events observed in the present. Such events shall help the planner draw conclusions that exceed the mere forecast provided. The concept of \textit{Event} can be found in different ontologies (\cite{ermolayev2008ontology,gottschalk2018eventkg}), having a prominent place in the YAMATO ontology (\cite{mizoguchi2010yamato}) as one of the core concepts.

The ontologies mentioned above provide a solid ground to create a unified ontology that serves the purpose of semantically enhancing explanations of demand forecasts issued by AI models.

\subsection{Methodologies for ontology design in manufacturing}\label{RELATED-WORK-MTD}
Multiple general methodologies were proposed to build an ontology. \cite{uschold1996building, uschold1995towards, fernandez1997methontology} suggest first identifying the purpose of the ontology, and \cite{uschold1996building,kim1995ontology} stress the importance of defining its scope and level of formality. Next,  \cite{kim1995ontology} suggests defining a problem statement and competency questions, which provide common ground to elicit knowledge from multiple sources, identify key concepts and relationships. When defining terms to be used in the ontology, an effort should be made to integrate pre-existing ontologies where possible.

Some authors identified steps specific to the creation of manufacturing domain ontologies. \cite{ameri2012systematic} developed a four-step methodology using a Simple Knowledge Organization System framework to develop a thesaurus of concepts, and then identify relevant classes and provide logical constraints and rules. A similar approach was developed by \cite{chang2010development}, with an emphasis on manufacturing design.

The methodologies we mentioned above provide us valuable guidance when developing our domain-specific ontology, aiming to integrate domain-specific knowledge regarding demand forecasting with the demand forecasting models, predictions, media events, external datasets, and explanations.

\section{The Proposed Architecture}\label{ARCHITECTURE}

We propose a modular architecture to provide forecast explanations. The main novelty is combining semantic technologies and media events to build the context and provide an informed prediction explanation.

\begin{figure*}[!htb]
\centering
\includegraphics[width=5.0in]{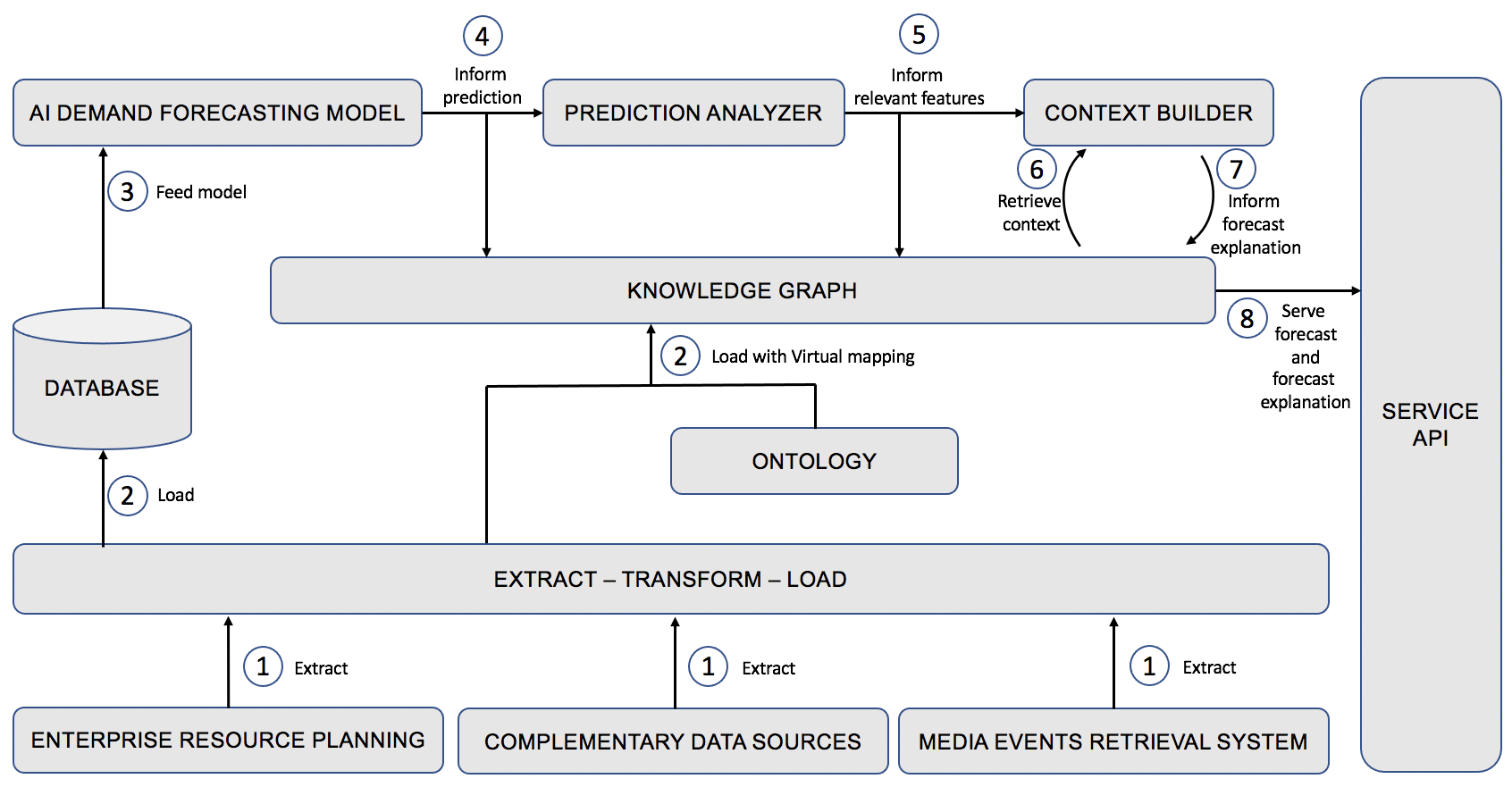}
\caption{Semantic XAI architecture for demand forecasting.}
\label{F:XAI-ARCHITECTURE}
\end{figure*}

\subsection{Architecture overview}\label{ARCHITECTURE-OVERVIEW}
The proposed supporting architecture enables us to realize the explanations for each prediction. The architecture integrates predictions, gathers insights on relevant features, incorporates domain knowledge and context to each prediction, and provides a forecast explanation to the end-user. The architecture (see Fig.~\ref{F:XAI-ARCHITECTURE}) comprises the following components:

\begin{itemize}
  \item \textbf{Enterprise Resource Planning}: software used by the manufacturing company to keep track of operations at different levels. The specific use case of demand forecasting may provide data regarding strategic sales planning, past and open sales, buyers risk assessment, and past demand.
  
  \item \textbf{Complementary data sources}: sources of data that provide valuable complementary data to train demand forecasting models.
  
  \item \textbf{Media Event Retrieval System}: is a system that keeps track of events reported in the media. Media Event Retrieval System uses Natural Language Processing to identify entities and concepts reported in media news and identify which news report about the same events. Wikification is applied to identify concepts, cluster news, and link events. 
  
  \item \textbf{Ontology}: encodes domain knowledge regarding demand forecasting, demand forecasting models, events retrieved from the Media Event Retrieval System, and available datasets. The Extract-Transform-Load module uses it to guide a virtual mapping procedure and instantiate the Knowledge Graph.
  
  \item \textbf{Knowledge Graph}: is the instantiation of the ontology, based on data provided by the Extract-Transform-Load module, and outcomes informed by multiple architecture components. The Knowledge Graph keeps track of data up to a configurable time window.
  
  \item \textbf{Extract-Transform-Load}: provides means to interface with the Enterprise Resource Planning and Media Events Retrieval System to retrieve relevant data, transform it as required, and ingest it to a database or into the Knowledge Graph. A virtual mapping procedure specifies how data pieces are related to concepts defined in the ontology and how it should be instantiated to the Knowledge Graph. The Extract-Transform-Load module consists of a series of batch processes. The processes are executed regularly to ensure the KG and database information are up to date regarding the Enterprise Resource Planning software and Media Event Retrieval System.
  
  \item \textbf{Database}: stores data relevant to the AI models, which can be used to train them.
  
  \item \textbf{AI Demand Forecasting model}: such a model can be built with a variety of algorithms, depending on the demand type and data available. Such models can be machine learning regression (\cite{sharma2012sales,bruhl2009sales}), deep learning (\cite{babai2020empirical,lolli2017single}) or probabilistic models (\cite{salinas2020deepar}). Demand forecasting models shall provide a demand estimate for future points in time together with some uncertainty estimation. This information is stored in the Knowledge Graph.
  
  \item \textbf{Prediction Analyzer}: interfaces with the \textbf{AI Demand Forecasting model} to determine which features are most important to a given prediction. To that end, it may use different black-box explainability techniques to assess relevant features under different criteria. The analysis outcomes are stored in the Knowledge Graph.
  
  \item \textbf{Context Builder}: has the responsibility to gather required pieces of data to issue good explanations. Interfaces with the Prediction Analyzer to get relevant features to each prediction. With the Knowledge Graph to obtain context regarding feature values, related events reported in the media, and complementary data that may be used to enhance the model. The Knowledge Graph also provides information regarding high-level concepts that describe the features (see Fig.~\ref{F:FEATURE-VECTOR-HIERARCHY}). This are used to inform on relevant features at a higher abstraction level, to avoid exposing model features and enable more concise explanations.
  
  \item \textbf{Service API}: serves predictions and their explanations to the end-users. It hides explanation details based on the end-user profile to expose only facts that are relevant to the user.
\end{itemize}

\subsection{Context building}\label{ARCHITECTURE-CB}
The goal of the architecture is to support semantically enhanced explanations for demand forecasting models. A key role is played by the \textit{Prediction Analyzer} and \textit{Context Builder} components. The \textit{Prediction Analyzer} provides an abstraction layer on top of black-box explainability models to inform on features that are most important to certain prediction. As mentioned in Section \ref{RELATED-WORK-BBM}, good explanations should convey useful information, target a specific user profile, and focus on actionability. In this line, the \textit{Context Builder} holds the responsibility to provide explanations that (I) convey the predicted demand and uncertainty, (II) provide insights on what concepts were considered to create the prediction, (III) provide relevant context, and (IV) target a specific user. The predicted value and the associated uncertainty are queried from the Knowledge Graph.

\begin{figure*}[!t]
\centering
\includegraphics[width=5.0in]{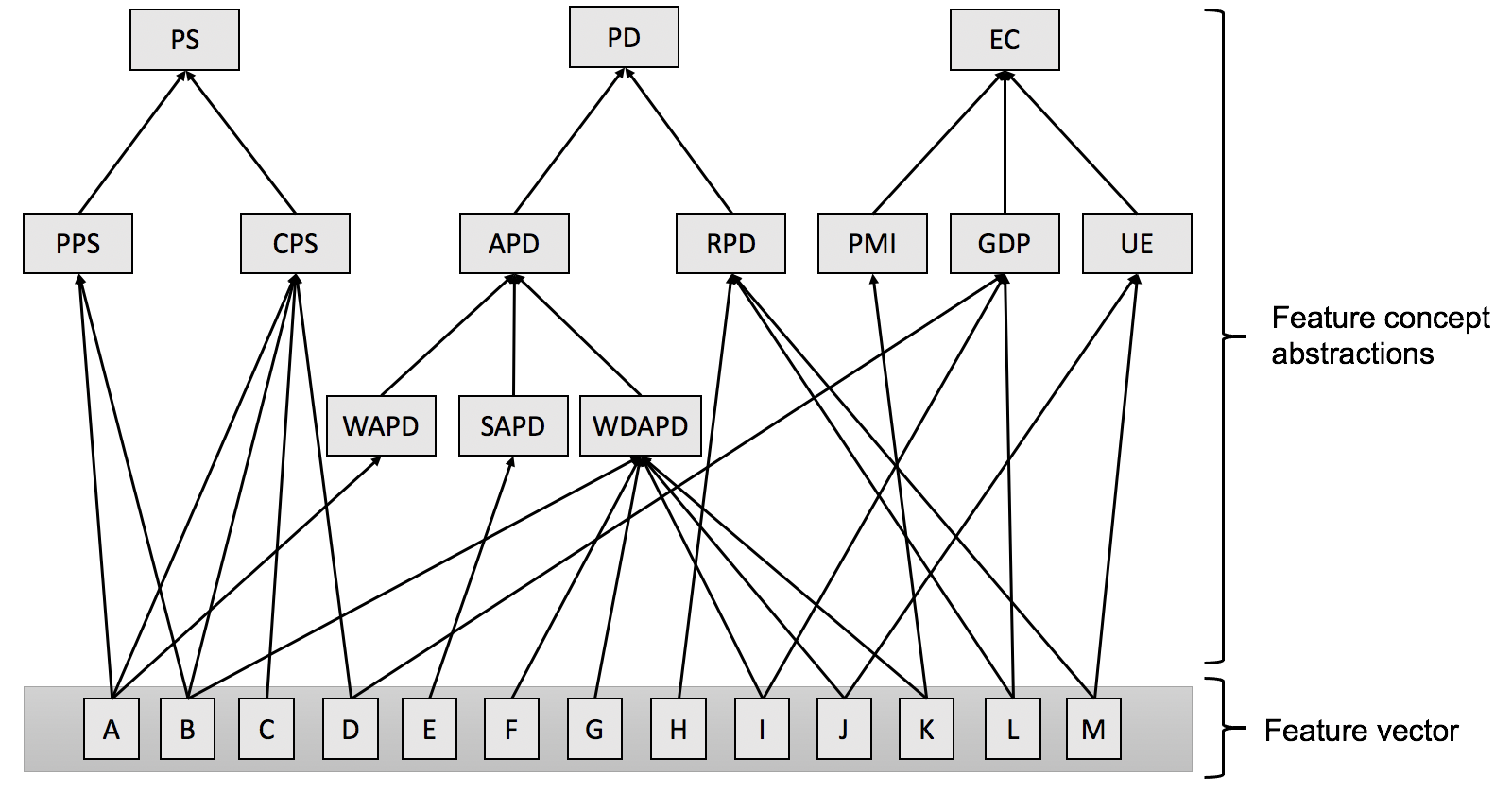}
\caption{Feature vector attributes and high-level concepts hierarchy associated with them. Considered attribute abstractions are PS (Planned Sales), PPS (Past Planned Sales), CPS (Current Planned Sales), PD (Past Demand), APD (Adjusted PD), SAPD (Scaled APD), WAPD (Weighted APD), WDAPD (Working Day APD), RPD (Raw Past Demand), EC (Economic Context), PMI (Purchasing Managers' Index), GDP (Gross Domestic Product), UE (Unemployment Rate). Features A-M are defined in Table~\ref{T:FEATURE-VECTOR-REFERENCE}}
\label{F:FEATURE-VECTOR-HIERARCHY}
\end{figure*}

To realize (II), the \textit{Context Builder} interfaces with the \textit{Prediction Analyzer} and the \textit{KG} to obtain information regarding relevant features to certain prediction. These features are mapped to a hierarchy of attribute concept abstractions (see Fig.~\ref{F:FEATURE-VECTOR-HIERARCHY}). At least one level of abstraction is considered to avoid exposing sensitive information regarding features used in the demand forecasting model and provide more concise explanations. 

To realize (III), it is necessary to consider specific data used to build the feature vector, events related to them, and additional knowledge we may gain from them. To that end, we provide a set of keywords that describe each feature. The keywords are taken into account by the Extract-Transform-Load process to issue queries against the Media Event Retrieval System. To do so, we consider the point in time each feature in the feature vector refers to. Based on it, we associate related events that are temporally close to it. Natural Language Processing techniques are applied to retrieved media events to extract new keywords relevant to the demand forecasting problem. Those new keywords are used to query for external datasets that could enrich the existing demand forecasting model. Media events, keywords we extracted from them, and metadata regarding external datasets are persisted to the Knowledge Graph and linked to specific features from feature vector instances. 

Finally, (IV) is realized by providing a multi-part explanation, where components are displayed or hidden, based on what kind of information is relevant to the end-user or confidentiality policies. This responsibility is delegated to the \textit{Service API}. E.g., a target audience could be planners interested in the demand forecast, associated uncertainty, and some context regarding feature and media events related to them. Another target audience could be experts who develop demand forecasting models. They would benefit from the data mentioned above and the possibility to drill-down from described concepts to specific model features. Such an explanation could be enriched with insights on frequent keywords found in related events reported in the media and open datasets that may be useful to the model.

\begin{table*}[!ht]
\centering
\resizebox{\columnwidth}{!}{
\begin{tabular}{|l|c|l|l|}
\hline
Feature ID & Actionable? & Feature definition & MERS query keywords \\ \hline
A & \multirow{4}{*}{YES} & equation & \begin{tabular}{@{}l@{}}car sales demand \\new car sales \\vehicle sales \\car demand \\automotive industry\end{tabular} \\ \cline{1-1} \cline{3-4} 
B &  & equation & Same as A \\ \cline{1-1} \cline{3-4} 
C &  & equation & Same as A \\ \cline{1-1} \cline{3-4} 
D &  & equation & \begin{tabular}{@{}l@{}} global GDP projection \\global economic outlook \\economic forecast\\ + keywords listed in A\end{tabular} \\ \hline
E & \multirow{9}{*}{NO} & equation & Same as A \\ \cline{1-1} \cline{3-4} 
F &  & equation & Same as A \\ \cline{1-1} \cline{3-4} 
G &  & equation & Same as A \\ \cline{1-1} \cline{3-4} 
H &  & equation & Same as A \\ \cline{1-1} \cline{3-4} 
I &  & equation & Same as D \\ \cline{1-1} \cline{3-4} 
J &  & equation & \begin{tabular}{@{}l@{}}unemployment rate \\unemployment numbers \\unemployment report \\employment growth \\long-term unemployment \\+ keywords listed in A\end{tabular} \\ \cline{1-1} \cline{3-4} 
K &  & equation & \begin{tabular}{@{}l@{}}purchase managers' index \\+ keywords listed in A\end{tabular} \\ \cline{1-1} \cline{3-4} 
L &  & equation & Same as D \\ \cline{1-1} \cline{3-4} 
M &  & equation & Same as J \\ \hline
\end{tabular}
}
\caption{Sample feature vector and keywords considered for the use case. We use the following abbreviations: sp (sales plan), GDP (Gross Domestic Product), PMI (Purchasing Managers' Index), wdp (monthly demand averaged per amount of working days in that month), demand (demand in a given month), lagNm (N months lag), pastwavg (weighted average for a given month in past years), scaled (value scaled between 0-1) \label{T:FEATURE-VECTOR-REFERENCE}}
\end{table*}

\section{Ontology}\label{ONTOLOGY}
Following related work elaborated in Section \ref{RELATED-WORK-DSO}, we created an ontology that describes concepts related to demand forecasting. Specifically, it describes the use of data and AI models to predict demand, demand forecasts, and forecast explanations. The forecast explanations provide context regarding feature relevance, potentially related media events, and external datasets that could be used to enrich future models. The ontology was published and is available online (\cite{DVN/GIREAY_2020}).

Upper ontologies provide a guide on how to think about the target domain when building a domain ontology. We use BFO as our upper ontology, which supports a world's snapshot view. Our entities can be of two types: occurrent (entities that occur) or continuent (entities that persist in time).

Regarding manufacturing, we reused the concepts of \textit{Product}, and \textit{Event} from the ontologies developed by \cite{leitao2006adacor}, \cite{borgo2007foundations}, and \cite{kourtis2019rule}. In \cite{cannataro2003data} and \cite{panov2014ontology}, we found the concepts of \textit{Dataset Specification}, \textit{Dataset}, \textit{Algorithm}, \textit{Datamining Algorithm}, \textit{Regression Algorithm}, \textit{Predictive Model}, \textit{Regression Model}, and \textit{Prediction}. \cite{gottschalk2018eventkg} provided us the concept of \textit{Information Provenance} to represent data sources. Finally, we introduced the following concepts: 
\begin{itemize}
	\item \textit{Media Reported Event}: relates to some \textit{Event} that was reported in news media;
	\item \textit{Media Reported Event Keyword}: keyword obtained from some \textit{Media Reported Event};
	\item \textit{External Dataset Metadata}: a piece of metadata describing some external dataset;
	\item \textit{Forecast Explanation}: a description providing reasons for a certain forecast. Relates to a specific \textit{Prediction}, \textit{Media Reported Event}, and \textit{External Dataset Metadata};
	\item \textit{Attribute}: a quantity describing an instance;
	\item \textit{Attribute abstraction}: a high-level concept describing some aspect of a dataset attribute;
	\item \textit{Feature Vector}: a vector containing attribute values describing certain an instance of data;
\end{itemize}

\section{Case Study}\label{USE-CASE}
Among the related research presented in Section \ref{RELATED-WORK}, we did not find scientific contributions focused on the explainability of AI demand forecasting models. The architecture we present in this research aims to provide the main building blocks required to provide adequate explanations for demand forecasts at a local level. To build and test the architecture, we worked with real-world data related to the automotive industry, provided by partners from FACTLOG and STAR European Horizon 2020 projects. Data was provided for 56 different materials. We considered a window of three years of data to build demand forecasting models for them. We issued predictions and built explanations for all materials in the last three months of data. Our demand forecasting models use the Support Vector Regression algorithm (\cite{drucker1997support}). We used nested cross-validation (\cite{stone1974cross}) to evaluate the models and make sure to use the latest data available for each target month. Finally, we used LIME (\cite{ribeiro2016should}) to implement the \textit{Prediction Analyzer}.

The \textit{Prediction Analyzer} provides a ranking of features for each forecast. For each of those features, we consider the query keywords presented in Table~\ref{T:FEATURE-VECTOR-REFERENCE} to query the Media Events Retrieval System and retrieve media events that provide context to feature values. We used Event Registry (\cite{leban2014event}), a well-established Media Events Retrieval System that monitors mainstream media since 2014. Our queries consider the keywords that describe each feature and the point-in-time in which each feature component took place. E.g., suppose a feature considers (I) GDP values three and fifteen months before the forecasting horizon and (II) demand four months before the forecasting horizon. In that case, the query will ask (I) for media events that relate to GDP at three and fifteen months before the forecasting horizon, and (II) media events regarding demand four months before the forecasting horizon.

We considered only media events in the English language. Once obtained, we processed them to identify and lemmatize nouns that provide insights on the most important facts related to the features. We also use these nouns to query for open datasets that may be used to enrich the current model. In particular, we queried the EU Open Data Portal\footnote{https://data.europa.eu}, a portal that provides access to open data published by EU institutions and bodies. For each matching external dataset metadata entry (provided or translated to the English language), we processed their title and description to create a word embedding with the Word2Vec algorithm (\cite{mikolov2013distributed}), using Google News 300 pre-trained model\footnote{https://code.google.com/archive/p/word2vec}. We create another embedding with the feature keywords from Table~\ref{T:FEATURE-VECTOR-REFERENCE}, and the keywords obtained from the media events we retrieved. To rank datasets metadata relevance, we compute the word movers distance (\cite{kusner2015word}) between both embeddings. Finally, to increase the diversity of displayed media event entries and datasets' metadata, we random sample them among top candidates for each case.

To create the ontology, we used Protege (\cite{noy2003protege}). The ontology served as a guideline to implement virtual mapping functions, to map data to the knowledge graph. We implemented the knowledge graph in Neo4j\footnote{https://neo4j.com}.

\begin{figure*}[!htb]
\centering
\includegraphics[width=5.0in]{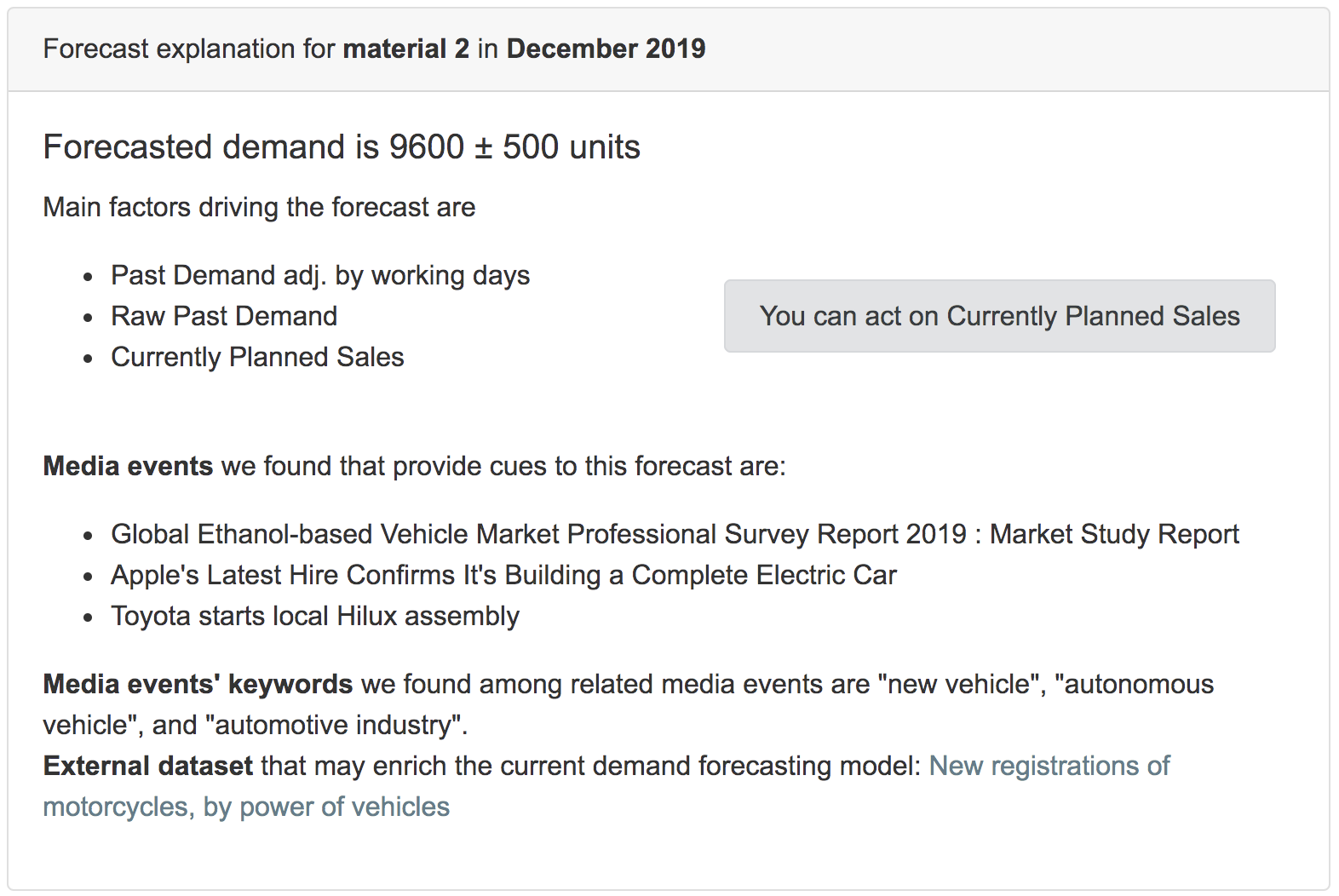}
\caption{An example of a demand forecast explanation that is displayed to the end user based on data retrieved from the \textit{Service API}.}
\label{F:XAI-SAMPLE-EXPLANATION-ENTRY}
\end{figure*}

We present a sample demand forecast explanation in Fig.~\ref{F:XAI-SAMPLE-EXPLANATION-ENTRY}. The explanations display the following information:
\begin{itemize}
 \item forecasted demand and associated uncertainty, for a given material at a certain point in time;
 \item main factors driving the forecast, which are obtained considering most important features to the forecast as provided by the \textit{Prediction Analyzer}, and abstracted using domain knowledge from the \textit{Knowledge Graph};
 \item a highlight regarding an actionable aspect, if it exists;
 \item media events associated with the most important features of the particular forecast instance;
 \item media events' keywords frequently found in media events related to the forecast instance;
 \item external dataset that may be used to enrich the existing demand forecasting model.
\end{itemize}

\section{Evaluation}\label{EVALUATION}
In our research, we develop an architecture that aims to provide explanations for demand forecasting. Since media events play an essential role in our forecast explanations, we were interested in assessing (A) if media events describe features context and (B) if retrieved external datasets were related to the features for which we queried them. Regarding the explanations, we assessed metrics related to listed external datasets, events, and keywords to understand if the provided entries were accurate and diverse. To measure diversity, we computed the ratio of diverse entries (RDE) defined in Eq.~\ref{E:RDE}. It is important to note that media events, keywords, and external datasets listed on explanations do not repeat themselves in a single explanation. However, the same media event, keyword, or external dataset reference may be found in another explanation. The best possible case would be to find completely different but accurate media events, keywords, and external datasets across all explanations. Such a scenario would maximize end-users learning.

In particular, for listed external datasets, we computed the accuracy and RDE. At the same time, for media events and related media events keywords, we measured the average precision@K and RDE@K with K=1,3 over all 168 forecast explanations. We selected K=1,3 since three media events and media events' keywords are listed in each explanation. Average precision@K is computed as the average of the precision@K we computed for each forecast explanation. RDE@K is computed as RDE across all 168 explanation entries, considering only the first K listed media event or media events' keywords entries.

\begin{equation}\label{E:RDE}
    RDE = \frac{\textit{Unique Entries}}{\textit{Total Listed Entries}}
\end{equation}

We evaluated these aspects by manually annotating a random sample of retrieved media events, retrieved external datasets metadata, and all media events, keywords, and external datasets proposed in demand forecast explanations.

We retrieved 128226 media events related to the features of interest and manually annotated a random sample of 528 media events. Regarding (A), we found that 69\% of retrieved media events were meaningful to features explaining a specific forecast. All media events were processed to search and retrieve information regarding external datasets, as described in Section~\ref{USE-CASE}. From the 401 external dataset entries we retrieved and annotated, we found that (B) 64\% were relevant to our use case.

We manually annotated 168 forecast explanations collected over three months. Each forecast explanation listed three media events, three media events keywords, and an external dataset recommendation. We manually annotated all of them. Results of their evaluation are presented in Table~\ref{T:RESULTS}.

\begin{table*}[!h]
\centering
\resizebox{\columnwidth}{!}{
\begin{tabular}{|l|l|l|}
\hline
\textbf{} & \textbf{Metric} & \textbf{Value} \\ \hline

\multirow{4}{*}{\textbf{Media Events}} & average precision@1 & 0.97 \\ \cline{2-3} 
 & average precision@3 & 0.97 \\ \cline{2-3} 
 & RDE@1 & 0.30 \\ \cline{2-3} 
 & RDE@3 & 0.11 \\ \hline
\multirow{4}{*}{\textbf{Media Events' Keywords}} & average precision@1 & 0.77 \\ \cline{2-3} 
 & average precision@3 & 0.78 \\ \cline{2-3} 
 & RDE@1 & 0.14 \\ \cline{2-3} 
 & RDE@3 & 0.09 \\ \hline
\multirow{2}{*}{\textbf{External Datasets}} & accuracy & 0.56 \\ \cline{2-3} 
 & RDE & 0.41 \\ \hline
 \end{tabular}}

\caption{Results we obtained from the analysis of forecast explanations issued for 56 products over three months. Media Events, Media Events' Keywords, and External Datasets correspond to contextual information displayed for each forecast explanation.\label{T:RESULTS}}

\end{table*}

We compiled relevant features to each forecast, associated media events, their annotated subset, and the annotated suggested datasets into a dataset made publicly available at \cite{DVN/FPXYCM_2020}.

When annotating the explanations, we found some interesting cases. One of them was the suggestion of ozone pollution dataset to enrich the automotive demand forecasting model. After some research, we found that while ozone is not emitted directly by automobiles, it is formed in the atmosphere due to a complex set of chemical reactions involving hydrocarbons, oxides of nitrogen, and sunlight. Another interesting case was a news media entry describing wireless charging. This media event seemed unrelated to our use case, but we confirmed that electric vehicles' demand drives research and development of a segment of wireless charging solutions. Finally, the most frequently listed media event among our forecast explanations was \textit{"Number of people in work surges again to a new record high of 32.3m"}. We consider this event relates to the economic context and is relevant to our particular demand forecasting case.

\section{Conclusion}\label{CONCLUSION}
In this research work, we presented an architecture that supports building semantically enhanced explanations for AI models' demand forecasts. We evaluated the architecture and explanations on a real-world use case developed with data obtained from European Horizon 2020 projects FACTLOG and STAR.

Current explanations provide demand forecast values, the associated uncertainty, high-level description of features that influenced the prediction, related media events, and a reference to some external dataset, providing context and means to improve the demand forecasting model. The high-level description of features informs main factors influencing the forecast, avoid exposing sensitive details regarding the demand forecasting models, and help to identify potential model bias.

Our future work will focus on further pre-processing media events, media events' keywords, and external datasets, to filter out those that are not relevant to the demand forecasting context and increase the forecast's explanation quality. At least three additional research directions can be pursued. First, explore means to increase the diversity of recommended media events and associated keywords displayed on forecast explanations. Second, evaluate the impact of incorporating suggested external datasets to the performance of existing demand forecasting models. Finally, we envision the explanations can be enhanced by including meaningful current events reported by the media. This way, in addition to a good understanding of the past context, the explanations will provide information on events that are likely to influence future demand so that the user can gain a better judgment and perspective when making a decision.

\section*{Acknowledgement}
This work was supported by the Slovenian Research Agency and the European Union’s Horizon 2020 program projects FACTLOG under grant agreement H2020-869951, and STAR under grant agreement number H2020-956573.





\bibliographystyle{elsarticle-harv}
\bibliography{bibliography}

\end{document}